\documentclass{article}
\usepackage{ijcai25}

\usepackage{times}
\usepackage{soul}
\usepackage{url}
\usepackage[hidelinks]{hyperref}
\usepackage[utf8]{inputenc}
\usepackage[small]{caption}
\usepackage{graphicx}
\usepackage{amsmath}
\usepackage{amsthm}
\usepackage{booktabs}
\usepackage{algorithm}
\usepackage[switch]{lineno}
\usepackage{algpseudocode}
\usepackage{float}

\usepackage{amssymb}

\newtheorem{definition}{Definition}

\title{Fast Explanations via Policy Gradient-Optimized Explainer}

\author{
Deng Pan
\and
Nuno Moniz\and
Nitesh V. Chawla
\affiliations
Lucy Family Institute for Data \& Society, University of Notre Dame\\
    Notre Dame, IN 46556 USA \\
\emails
\{dpan, nuno.moniz, nchawla\}@nd.edu
}

\begin{document}
\maketitle

\begin{abstract}
The challenge of delivering efficient explanations is a critical barrier that prevents the adoption of model explanations in real-world applications. Existing approaches often depend on extensive model queries for sample-level explanations or rely on expert's knowledge of specific model structures that trade general applicability for efficiency. To address these limitations, this paper introduces a novel framework Fast EXplanation (FEX) that represents attribution-based explanations via probability distributions, which are optimized by leveraging the policy gradient method. The proposed framework offers a robust, scalable solution for real-time, large-scale model explanations, bridging the gap between efficiency and applicability.
We validate our framework on image and text classification tasks and the experiments demonstrate that our method reduces inference time by over 97 percent and memory usage by 70 percent compared to traditional model-agnostic approaches while maintaining high-quality explanations and broad applicability. 

\end{abstract}

\section{Introduction}
While deep classification models demonstrate superior performance across a range of tasks, their "black-box" nature often hinders their acceptance and deployment in critical areas such as healthcare, finance, and autonomous systems \cite{miotto2018deep,ozbayoglu2020deep,grigorescu2020survey}. In these high-stakes contexts, it is essential not only to achieve high predictive accuracy but also to provide clear, understandable explanations of the models' decisions to foster trust and ensure accountability.

Despite the progress in explainable AI (XAI) research \cite{atakishiyev2021explainable,singh2020explainable}, achieving explainability in real-world, large-scale applications remains challenging. A significant barrier is the difficulty of providing \emph{efficient explanations} that can scale without imposing prohibitive computational costs. Current attribution-based explanation methods often require substantial computation during inference, making them impractical for time-sensitive tasks and large-scale deployment \cite{chuang2023efficient,lundberg2020local}. Therefore, improving the efficiency of explanations is critical for enabling their broader adoption in real-world applications.

In practice, when working with black-box models or complex architectures, it is expected to use \emph{model-agnostic explanation methods} \cite{ribeiro2016should,lundberg2017unified,petsiuk2018rise,fong2017interpretable}. These methods have the advantage of being applicable to a wide range of models, but they often require numerous additional forward passes or gradient computations, making them inefficient and costly for real-world applications.

In scenarios where we have full access to the model’s architecture, such as CNNs or Transformers, \emph{model-specific explanation methods} can be employed to provide rapid explanations  \cite{selvaraju2017grad,chefer2021transformer,qiang2022attcat}. These methods are tailored to specific model structures, leveraging the unique behaviors of certain architectures to achieve efficient explanations. 
However, in real-world settings, models are often either black-box or not easily categorized into standard architectures, which limits the application of model-specific explanations. 

To address the inefficiency of model-agnostic methods and the limitation of model-specific approaches, \emph{amortized explanation} techniques have been proposed by training a deep neural network (DNN) to approximate an explanation distribution, thereby accelerating model-agnostic explanations to a single forward pass during inference \cite{chuang2023efficient,jethani2021fastshap,chen2018learning}.  However, these methods rely heavily on approximating specific proxy explanation methods, such as SHAP, by treating their explanations as ground truth (or pseudo-label). This introduces limitations: the performance of amortized methods is inherently capped by the quality of the proxy explanations; and they also rely on the assumptions made by the proxy methods.


{\it{In this work}}, we propose a novel policy gradient based approach to learn a model-specific explainer that is not only capable of making fast explanation to any black-box models, but also have no reliance on pseudo-labels from existing proxy explanation methods.

Our {\bf{main contributions}} are summarized as follows: 1) To the best of our knowledge, this is one of the first work that leverages reinforcement learning to directly learn an efficient explainer directly from data and the prediction model. 2) Unlike other amortized methods, our method doesn't rely on the pseudo-labels provided by any proxy explanation method, such as SHAP. 3) A KL-divergence regularization is also introduced to enhance the generalizability of the learned explainer.   4) Comprehensive qualitative and quantitative experiments across multiple datasets demonstrate the superior quality and efficiency of our approach.

\section{Related Work}

In this work, we focus on explanations in the format of feature attribution \cite{linardatos2020explainable}, i.e., finding the importance score for individual input features that influence a prediction.
Therefore, we review three categories of feature attribution methods that are closely related to our approach:\textit{ model-agnostic approaches}, \textit{model-specific approaches}, and  \textit{amortized approaches}.

   \paragraph{Model-agnostic approaches} Model-agnostic approaches are designed to be broadly applicable, making minimal or no assumptions about the to-be-explained prediction models. One common strategy involves using an explainable surrogates to approximate the local behavior of models, which is particularly useful for black-box models. For instance, LIME \cite{ribeiro2016should} fits a surrogate interpretable model (such as a linear model) to explain predictions locally by perturbing the input data and observing the changes in predictions. Similarly, SHAP \cite{lundberg2017unified} leverages Shapley values from game theory to ensure a unique surrogate solution with desirable properties such as local accuracy, missingness, and consistency. RISE \cite{petsiuk2018rise} generates saliency maps by sampling randomized masks and evaluating their impact on the model's output. 

 Another category of model-agnostic techniques leverages gradient information from white-box models to provide explanations. Instead of learning surrogates, these methods exploit locally smoothed gradients to approximate the model's local behavior. The smoothing strategies vary among approaches. For instance, Integrated Gradients \cite{sundararajan2017axiomatic} computes explanations by averaging gradients of interpolated samples between a baseline input and the target input. AGI \cite{pan2021explaining} refines this concept by averaging gradients along multiple adversarial attack trajectories, while NeFLAG \cite{li2023negative} utilizes gradients averaged over a hyperspherical neighborhood. 

 Although these methods are usually widely applicable, they are resource-intensive during inference due to the necessity of a large number of additional model queries.
   
   \paragraph{Model-specific approaches} Model-specific approaches are typically tailored to specific model architectures, enabling efficient explanations by utilizing attention weights, convolutional feature maps or custom layers. GradCAM \cite{selvaraju2017grad}, for instance,  uses the weighted average of the convolutional feature maps to generate attributions, effectively working on CNNs. Similarly, methods like AttLRP \cite{chefer2021transformer} and AttCAT \cite{qiang2022attcat} are designed specifically for transformer-based models, relying on attention weights from various attention heads and layers to compute final explanations.  DeepLIFT \cite{shrikumar2017learning} provides a framework for explaining deep learning models under the condition that propagation rules can be adapted. 

   \paragraph{Amortized approaches} Amortized explanation methods approximate the explanations from the resource-heavy model-agnostic methods (proxy methods) by a single forward pass. For example, FastSHAP \cite{jethani2021fastshap}, which amortizes the cost of fitting kernelSHAP by stochastically training a neural network to approximate it globally. CoRTX\cite{chuang2023cortx}, on the other hand, learns the explanation-oriented representation in a self-supervised manner and reduces the dependence of training on pseudo-labels from proxy methods. Overall, this type of methods achieves efficiency via an additional global surrogate function on top of the surrogates in model-agnostic methods, which introduces additional uncertainty.

In this paper, we propose a novel reinforcement learning framework that learns a distribution-based explainer that achieves the universality of model-agnostic approaches and the efficiency of model-specific approaches without relying on any proxy methods.

\section{Proposed Method: Fast Explanation (FEX)} \label{sec:proposed}
\begin{figure*}[th]
    \centering
    \includegraphics[width=0.9\linewidth]{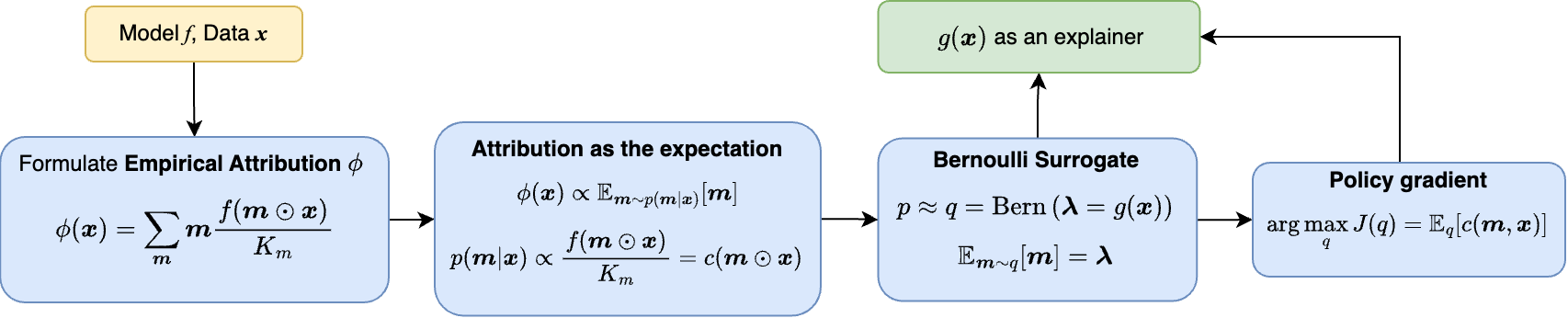}
    \caption{An illustration of the proposed method. The process begins with  empirical attribution, calculated by summing over $2^N$ terms. To address the computational intractability of this summation, the attribution is reformulated as an expectation over a probability distribution $p$. Subsequently, $p$ is approximated by a Bernoulli distribution $q$, enabling a closed-form solution that depends solely on the parameters of $q$. Finally, the parameters of $q$ are optimized using the policy gradient method, yielding an approximation of the empirical attribution.}
    \label{fig:diagram}
\end{figure*}

We lead with a discussion of an intractable empirical attribution, which relies on an exhaustive search over all possible feature combinations. Then, we interpret this empirical attribution as an expectation of a probability distribution and approximate this distribution via a policy gradient approach. Figure~\ref{fig:diagram} illustrates the proposed model.

\subsection{Empirical Attribution}

For an input comprising $N$ features, there exists $2^N$ distinct feature selection combinations. We represent each feature combination using a mask $\boldsymbol{m}\in \{0,1\}^N$. \textit{For clarity, we define a mask entry of 0 to indicate that a feature is masked (removed), while a mask entry of 1 denotes that a feature is retained. }

Consider an input vector $\boldsymbol{x} = (x_1, ..., x_N)^\top$, with a \textit{ binary classification} function $f: X \to [0,1]$. For each masked version of the input, represented as $\boldsymbol{m}\odot \boldsymbol{x}$, we obtain a corresponding prediction $f(\boldsymbol{m}\odot \boldsymbol{x})\in [0,1]$.

Intuitively, the \textbf{native contribution} of a feature $x_i$ to $f(\boldsymbol{m}\odot \boldsymbol{x})$ can be represented by the average contribution from all features present in $\boldsymbol{m}$, i.e.,
\begin{align}
    c_i = \frac{f(\boldsymbol{m} \odot \boldsymbol{x}) }{K_{m}} 
\end{align}
where $K_{m}$ denotes the number of non-zero entries in $\boldsymbol{m}$, representing the number of retained features in the masked input. This intuition for naive contribution comes from the philosophy that \textit{all features should be viewed as equally important if there isn't any prior knowledge}. 

Assume that there is a subset $M_{i} = \{\boldsymbol{m}|\boldsymbol{m}\in \{0,1\}^N, m_i=1\}$, which comprises all masks that retain feature $x_i$. Then the sum of naive contributions of $x_i$ over $M_{i}$ is a good indicator of the importance of feature $x_i$. 

Building on this, we define the empirical attribution as the total naive contribution from all masked inputs that retain the feature $x_i$. Specifically:
\begin{definition}
    \label{def:emp_attri}
(Empirical Attribution) Given an input vector $\boldsymbol{x}\in X$,  a set of masks $M = \{0,1\}^N$, and the prediction function $f(\boldsymbol{x}): X \to [0, 1]$, the empirical attribution of a feature $x_i$ is defined as:
\begin{align}
    \phi_i(\boldsymbol{x}) 
    = \sum_{\boldsymbol{m}\in M_{i}} \frac{f(\boldsymbol{m}\odot \boldsymbol{x})}{K_m} =\sum_{\boldsymbol{m} \in M} m_i\cdot \frac{f(\boldsymbol{m}\odot \boldsymbol{x})}{K_m},
\end{align}
where $M_{i}=\{\boldsymbol{m}\in  M| m_i=1\}$ represents the set of masks retain feature $x_i$,  and $K_m$ represents the number of non-zero elements in $\boldsymbol{m}$.
\end{definition}

The cumulative empirical attribution can also be expressed in vector form when denoting $\phi(\boldsymbol{x})$ by:
\begin{align}
    \phi(\boldsymbol{x}) &= (\phi_1(\boldsymbol{x}), \phi_2( \boldsymbol{x}), ...,\phi_N(\boldsymbol{x}))^\top\
    \end{align}
Since $\boldsymbol{m} = (m_1, m_2, ..., m_N)^T$, the vector form of the empirical attribution becomes:
\begin{align}\label{eq:empirical_attribution}
 \phi(\boldsymbol{x}) =   \sum_{\boldsymbol{m}} \boldsymbol{m} \cdot \frac{f(\boldsymbol{m}\odot \boldsymbol{x})}{K_m},
\end{align}
where $\phi(\boldsymbol{x})\in \mathcal{R}_+^N$.

\subsection{Attribution as an Expectation}
Calculating the empirical attribution is computationally prohibitive due to its exponential complexity of $O(2^N)$ . A common approach to address this challenge is to approximate it via Monte Carlo simulation,
as has been similarly demonstrated in the RISE \cite{petsiuk2018rise} framework. 
However, Monte Carlo methods remain computationally intensive, and the quality of the attribution is highly dependent on the number of simulation steps. 

Note that the empirical attribution $\phi(\boldsymbol{x})\in \mathcal{R}_+^N$ in Eq~\ref{eq:empirical_attribution} can be written as a form of probability expectation with an appropriate normalization factor $A(\boldsymbol{x})= \sum_{\boldsymbol{m}} \frac{f(\boldsymbol{m}\odot \boldsymbol{x})}{K_m}$. Specifically, we have:
\begin{align}\label{eq:cea_expectation}
    \phi(\boldsymbol{x}) \propto \mathbb{E}_{\boldsymbol{m} \sim p(\boldsymbol{m}| \boldsymbol{x})} [\boldsymbol{m}].
\end{align}
where the distribution $p(\boldsymbol{m}|\boldsymbol{x})$ is defined by
\begin{align}
    p(\boldsymbol{m}|\boldsymbol{x}) = \frac{f(\boldsymbol{m}\odot \boldsymbol{x})}{A(\boldsymbol{x})\cdot K_m}.
\end{align}
Although it is still intractable to directly calculate the expectations, it is possible to obtain the expectation as a closed form if $p(\boldsymbol{m}|\boldsymbol{x})$ follows some specific distribution families.

\subsection{Tractable Bernoulli Surrogate}
For the purpose of explanation, we chose a multivariate Bernoulli distribution for its natural alignment with the disentanglement of different features. It serves as a surrogate to $p(\boldsymbol{m}|\boldsymbol{x})$:
\begin{align}
    q = \operatorname{Bern}\left(\boldsymbol{\lambda}  = g(\boldsymbol{x})\right),
\end{align}
where $\boldsymbol{\lambda}\in[0,1]^N$ is the mean parameter of the Bernoulli distribution, parameterized by a neural network $g(\boldsymbol{x})$.

Its expectation has a closed form. Specifically, for $\boldsymbol{m}$ sampled from $q$, we have:
\begin{align}
    \mathbb{E}_{\boldsymbol{m}\sim q}[\boldsymbol{m}] = \boldsymbol{\lambda}.
\end{align}
This property is particularly advantageous because it allows the mean parameter $\boldsymbol{\lambda}=g(\boldsymbol{x})$ to represent the empirical attribution directly if $q$ can approximate $p$, as defined in Eq~\ref{eq:cea_expectation}. 

However,  $p(\boldsymbol{m}|\boldsymbol{x})$ is not directly computable, instead, we start by defining a score function $c(\boldsymbol{m}, \boldsymbol{x})$  as
\begin{align} \label{eq:score_func}
    c(\boldsymbol{m}, \boldsymbol{x}) = \frac{f(\boldsymbol{m}\odot \boldsymbol{x})}{ K_m} \propto p(\boldsymbol{m}|\boldsymbol{x}) ,
\end{align}
Note that the expectation is primarily influenced by regions of the probability distribution with high density. Therefore, to approximate $p$ with $q$, we need to optimize $q$ such that the high density region matches $p$.  Therefore, assume there are $T$ masks $\boldsymbol{m}_1, ..., \boldsymbol{m}_T$ sampled from $q$, we aim to maximize the following objective:
\begin{align}\label{obj:maskgen}
    \max_q J(q) = \mathbb{E}_{q} \left[\frac{1}{T}\sum_{t=1}^Tc(\boldsymbol{m}_t, \boldsymbol{x})\right]  .
\end{align}

\subsection{Policy Gradient}
In the reinforcement learning literature, objectives with a structure similar to Eq.~\ref{obj:maskgen} can be effectively optimized using policy gradient methods. The policy gradient framework is characterized by four fundamental components: states (of the environment), actions (by the agent), the policy (for generating actions), and the return (of a series of actions).

\subsubsection{Policy Gradients Adaptation}
To adapt our framework to the policy gradient methodology, it is crucial to establish a clear correspondence between the key concepts in our approach and those traditionally utilized in the policy gradient literature. This section provides a detailed mapping of these conceptual alignments.

Lets rephrase our problem as follows: Given the score function $c$, and original model input $\boldsymbol{x}$, we need to find a distribution $q$ such that it maximizes Eq~\ref{obj:maskgen}.

\paragraph{Input $\boldsymbol{x}$ as Static States:}
In policy gradient framework, a state $s_t$ represents the current situation or configuration of the environment with which the agent interacts. In our context, since an input sample doesn't change over the masking actions, we consider these samples as static states. Formally, $s_t = \boldsymbol{x}$, where $t = 0, 1, ..., T$.

\paragraph{Mask $\boldsymbol{m}$ as Actions}
    In our framework, applying masks to static input samples can be viewed as actions applied towards the states. i.e., $a_t = \boldsymbol{m}_t$.

\paragraph{Bernoulli Surrogate $q$ as the Policy}
    The policy in reinforcement learning generates actions. Similarly, in our context, the mask distribution $q$ can be viewed as the policy that generates the masks. Specifically, $\boldsymbol{m}_t \sim q$.
\paragraph{Return}
    Consequently, the weighted score function $\frac{1}{T}c(\boldsymbol{m},\boldsymbol{x})$ performs as the reward given an action $\boldsymbol{m}$ upon state $\boldsymbol{x}$. Furthermore, if we define $\tau$ as a trajectory of a mask sequence $\boldsymbol{m}_1,...,\boldsymbol{m}_T$, the return $R$ can be computed by
    \begin{align}\label{return}
        R(\tau)=\sum_{\boldsymbol{m}_t\in \tau} \frac{c(\boldsymbol{m}_t, \boldsymbol{x})}{T} 
    \end{align}
\paragraph{Objective Function}
 The objective function can be formally expressed as
\begin{align}\label{obj:mask_traj}
J(q) = \mathbb{E}_{\tau \sim q}[R(\tau)]
\end{align}
where $\tau$ is a trajectory sampled from $q$.

\subsubsection{Policy Gradient Formulation}
Considering the above terminology connections between our framework and the policy gradient method, the gradient of the objective in Eq~\ref{obj:mask_traj} can be expressed as:
\begin{align}\label{eq:obj2}
    \nabla J(q) = \mathbb{E}_{\tau \sim q} \left[{\sum_{t=1}^{T} \nabla_{q} \log q(\boldsymbol{m}_t| \boldsymbol{x}) A^{q}(\boldsymbol{x}, \boldsymbol{m})}\right].
\end{align}

The advantage function $A^{q}(\boldsymbol{x}, \boldsymbol{m})$ is the difference between Action-Value function (Q-function)  and  Value function (V-function). The Q-function is defined as the expected return with the first action (mask) being $\boldsymbol{m}$
\begin{align}
    T\cdot Q^{q}(\boldsymbol{x}, \boldsymbol{m}) &= c(\boldsymbol{m},\boldsymbol{x}) + \mathbb{E}_{\tau \sim q} \left[\sum_{\boldsymbol{m_t}\in \tau, t\geq 2} c(\boldsymbol{m}_t,\boldsymbol{x})\right]\\
    &= c(\boldsymbol{m},\boldsymbol{x}) + (T - 1)\cdot \mathbb{E}_{\boldsymbol{m}\sim q} \left[c(\boldsymbol{m}, \boldsymbol{x})\right],
\end{align}
where the factor $T$ is multiplied on both sides for conciseness. Similarly, the V-function is defined by the expected return if the first action $\boldsymbol{m}$ is sampled from $q$:
\begin{align}
    T\cdot V^q(\boldsymbol{x}) &= \mathbb{E}_{\boldsymbol{m}\sim q}\left[c(\boldsymbol{m},\boldsymbol{x})\right] + (T - 1)\cdot \mathbb{E}_{\boldsymbol{m}\sim q} \left[c(\boldsymbol{m}, \boldsymbol{x})\right]\\
    &= T\cdot \mathbb{E}_{\boldsymbol{m}\sim q} \left[c(\boldsymbol{m}, \boldsymbol{x})\right].
\end{align}
Therefore, the advantage function can be obtained by subtracting $V$ from $Q$ . Specifically,
\begin{align}\label{eq:advantage}
    A^q(\boldsymbol{x}, \boldsymbol{m}) = \frac{1}{T}\cdot (c(\boldsymbol{m}, \boldsymbol{x}) - V^q(\boldsymbol{x})).
\end{align}
Note that $V^q(\boldsymbol{x})$ can also be approximated by a neural network $v(\boldsymbol{x})$, which can be trained by minimizing the following loss function:
\begin{align}\label{eq:value_mse}
    L_v(v) = \mathbb{E}_{\tau\sim q}\left[ \sum_{t = 0}^{T}\frac{1}{T}(c(\boldsymbol{m}_t, \boldsymbol{x}) - v(\boldsymbol{x}))^2\right].
\end{align}

\paragraph{Proximal Policy Optimization:} Policy gradient methods may suffer the issue of performance collapse when the policy changes too much during a single update. Therefore, we facilitate the clip trick used in PPO (Proximal Policy Optimization)\cite{schulman2017proximal} that constrains the update within each step. Consequently, the gradient in Eq~\ref{eq:obj2} can be written by:
\begin{align}\label{eq:ppo}
    &\nabla_{q} J(q) = \nabla_q L_{ppo} = \nabla_{q} \mathbb{E}_{\tau \sim q}\sum_{t=0}^{T}  L(t),  \\
    L(t) &=\min\left(\frac{q(\boldsymbol{m}_t| \boldsymbol{x})}{q^{\ell}(\boldsymbol{m}_t| \boldsymbol{x})} A^{q}(\boldsymbol{m}, \boldsymbol{x}), C A^{q}(\boldsymbol{m}, \boldsymbol{x}) \right),\\
    C &= \operatorname{clip}\left(\frac{q(\boldsymbol{m}_t| \boldsymbol{x})}{q^{\ell}(\boldsymbol{m}_t| \boldsymbol{x})}, 1-\epsilon, 1+\epsilon\right)
\end{align}
where $q^{\ell}$ represents the policy from the last updating step $\ell$.
Additionally, an \textbf{entropy regularization} term $H(q)$ is also added to balance the \textit{exploration and the exploitation} during the reinforcement learning steps.

Combining the PPO objective, the entropy, and the MSE loss for $v(\boldsymbol{x})$, the objective function can then be written by
\begin{align}
    L =  L_{ppo} - \lambda_{en}H(q) + \lambda_v L_v,
\end{align}

\begin{table*}[ht]
\centering
\resizebox{0.7\textwidth}{!}{ 
\begin{tabular}{|l|c|c|c|c|c|c|c|} \hline 

 & \textbf{FEX}  &\textbf{FastSHAP} & \textbf{RISE} & \textbf{IG} & \textbf{GradSHAP} & \textbf{GradCAM} & \textbf{AttLRP} \\ \hline  
\# propagation & $O(1)$  &$O(1)$ & $O(K)$ & $O(K)$ & $O(K)$ & $O(1)$ & $O(1)$ \\ \hline  
\# backpropagation & 0  &0 & 0 & $O(K)$ & $O(1)$ & $O(1)$ & $O(1)$ \\ \hline  
 Requires training& \checkmark  &\checkmark & \texttimes & \texttimes & \texttimes & \texttimes & \texttimes \\ \hline 
Proxy independent& \checkmark &\texttimes& --& --& --& --&  --\\ \hline  
Model Agnostic & \checkmark  &\checkmark & \checkmark & \checkmark & \texttimes & \texttimes & \texttimes \\ \hline  
Blackbox & \checkmark  &\checkmark & \checkmark & \texttimes & \texttimes & \texttimes & \texttimes \\ \hline
\end{tabular}
}
\caption{Comparison of computational costs, capabilities and limitations across different explanation methods. Here, $K$ denotes the number of queries to the prediction model.}
\label{tab:comparison}
\end{table*}

\subsection{Generalizability}

Generalizability in our framework involves two key aspects: (1) generalization over the distribution of all samples and (2) generalization across different output classes. Both are crucial for creating robust explainers that go beyond individual input-prediction pairs.

\paragraph{Generalization Over Sample Distribution:}
Generalization across samples ensures the explainer $g(\boldsymbol{x})$ consistently provides meaningful explanations over a dataset $\boldsymbol{X}$. When trained on a diverse dataset, our framework allows effective adaptation to diverse inputs.

\paragraph{Generalization Over Class Distribution:}
In multi-class classification, the prediction function $\boldsymbol{f}$ outputs a probability vector $(f_1, \dots, f_K)^\top$ over $K$ classes, requiring $K$ explainers $\boldsymbol{g}_1, \dots, \boldsymbol{g}_K$. Intuitively, when $f_i$ dominates the predicted probabilities, the corresponding explainer $\boldsymbol{g}_i\in(0,1)$ should also have a dominant average score. To enforce this alignment, KL-divergence is used to match the average explainer scores with the predicted class probabilities:

\begin{align}\label{eq:kldiv}
   L_{\text{kl}} = \mathcal{D}_{\text{kl}}\left(\operatorname{Softmax}\left(\frac{\sum_{i=1}^N \log \boldsymbol{g}_i}{N}\right), \boldsymbol{f}\right).
\end{align}
Averaging the log values of the explainer scores ensures a more stable computation, as log probabilities are additive by nature. The softmax function is then applied to form a valid probability distribution. This approach guarantees consistency between the explainers and the classifier's output, facilitating robust and scalable explanations across classes. Consequently, the overall objective function for the policy gradient adaptation becomes:

\begin{align}\label{eq:final_obj}
    L =  L_{ppo} - \lambda_{en}H(q) + \lambda_v L_v + \lambda_{kl}L_{kl} ,
\end{align}

\begin{algorithm}[t]
\caption{PPO for Fast Explanations}
\begin{algorithmic}[1]
\setlength{\baselineskip}{12pt} 
\State \textbf{Input:} training samples set $X$, prediction function $\boldsymbol{f}=(f_1, ..., f_K)^T$, initial explainer network $\boldsymbol{g}=(\boldsymbol{g}_1, ...,\boldsymbol{g}_K)^T$, initial value network $\boldsymbol{v}=(v_1, ..., v_K)^T$, and hyperparameters $\lambda_{en}$, $\lambda_v$, $\lambda_{kl}$
\For{$i = 1, 2, \ldots$}
    \State Get a batch of input-output pairs $X_i \subset \{(\boldsymbol{x}, y)| \boldsymbol{x}\in X, y = \arg\max_k f_k(\boldsymbol{x})\}$
    \For{$\ell = 0, 1, 2, \ldots$}
        \State Collect a set of trajectories $\mathcal{D}_{\ell} = \{\tau_j\}$ by running policy $q = \operatorname{Bern}(\boldsymbol{g}_y(\boldsymbol{x}))$ for all $(\boldsymbol{x}, y)$ pairs.
        
        \State Compute the Advantage $A^q(\boldsymbol{m}_t, \boldsymbol{x})$ by Eq~\ref{eq:advantage}.
        
        \State Obtain the PPO-Clip objective $L_{ppo}$ by Eq~\ref{eq:ppo}, value network MSE loss $L_v$ by Eq~\ref{eq:value_mse},  KL-divergence $L_{kl}$ by Eq~\ref{eq:kldiv}, and entropy $H(q)$.

        \State Update the policy by minimizing the objective $L$ in Eq~\ref{eq:final_obj}
    \EndFor
\EndFor
\end{algorithmic}
\label{algo:ppo}
\end{algorithm}

\begin{figure*}[ht]
    \centering
    \includegraphics[width=0.8\textwidth]{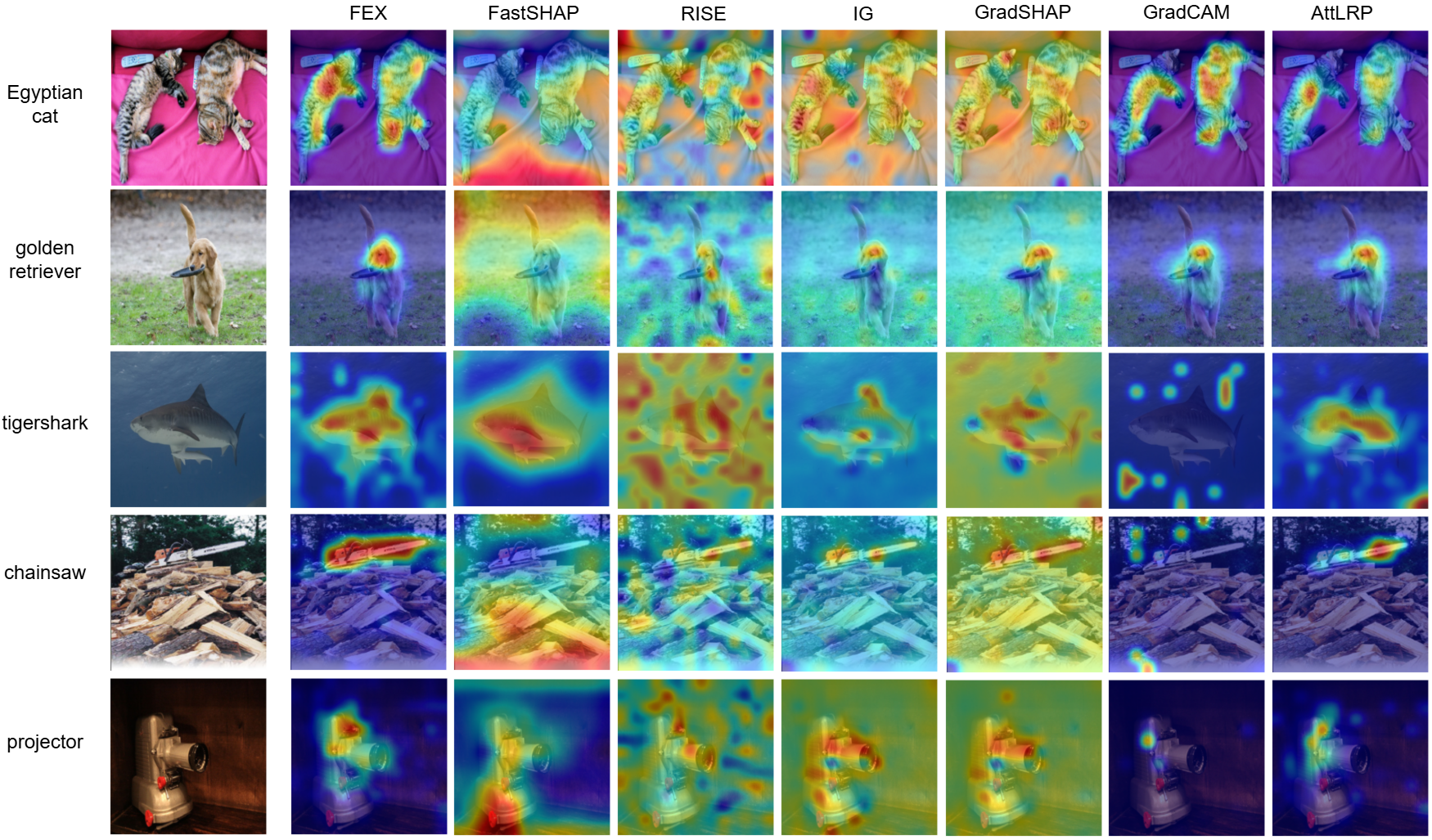}
    \caption{Qualitative examples for explaining the predictions in the image classification task. }
    \label{fig:cvexamples}
\end{figure*}

\subsection{Efficiency and Capabilities}
Table~\ref{tab:comparison} provides a detailed comparison of our Fast Explanation method (FEX) with several related explanation methods.

FEX distinguishes itself by requiring only $O(1)$ forward passes of $\boldsymbol{g}(\boldsymbol{x})$ during inference, ensuring exceptional computational efficiency. In contrast, other model-agnostic baselines, such as RISE, IG, and GradSHAP, require $O(K)$ queries to the prediction model. While methods like GradCAM and AttLRP achieve similar $O(1)$ efficiency in terms of model queries, they are inherently model-specific and therefore cannot be applied in a black-box setting.

Beyond its computational efficiency and model-agnostic nature, FEX offers an additional advantage: it does not depend on pseudo-labels generated by proxy explainers. For example, although FastSHAP achieves $O(1)$ efficiency, its reliance on pseudo-labels from SHAP introduces potential limitations, as its performance is constrained by the accuracy of the proxy explainer.

\begin{figure}[ht]
    \centering
    \includegraphics[width=0.8\linewidth]{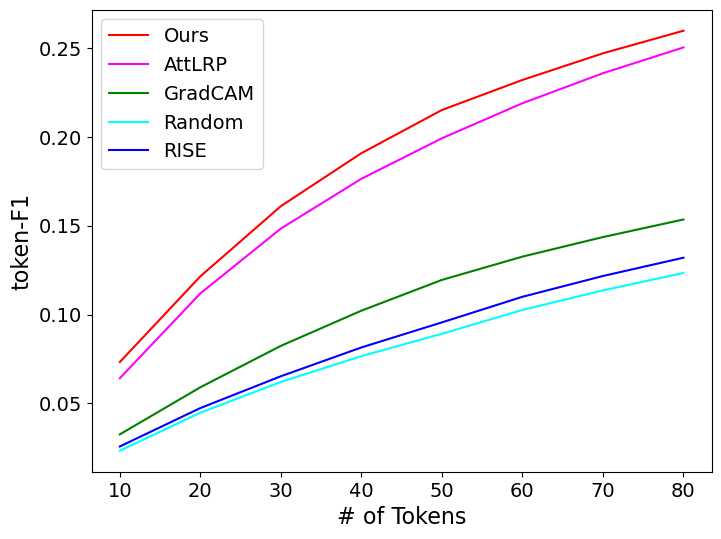}
    \caption{Quantitative evaluation results for the text classification task. The x-axis represents the number of text tokens inserted starting from the most important token, and the y-axis is the F1 score given that amount of tokens. The higher the better.}
    \label{fig:text_results}
\end{figure}

\section{Experiments} \label{sec:exp}
We conduct experiments on both image and text classification tasks. For image classification, we use the ViT model \cite{dosovitskiy2020image} fine-tuned on the ImageNet dataset \cite{deng2009imagenet} as the prediction model. The FEX explainer is finetuned on the full ImageNet dataset with 1.3M samples (FEX-1.3M) or a subset of 50,000 samples (FEX-50k) for one epoch . For text classification, we use the BERT model \cite{devlin2018bert} fine-tuned on the SST2 dataset \cite{socher2013recursive} for sentiment analysis. The FEX explainer is finetuned on the Movies Reviews \cite{zaidan2008modeling} dataset for one epoch with batch size 256. Unless otherwise specified, in all experiments, the $g(\boldsymbol{x})$ is set to the same architecture as the predictor $f$, with appended MLP prediction heads, and the hyperparameters are set to $\lambda_{en}=10^{-5}$, $\lambda_{v}=0.5$ and $\lambda_{kl}=1$.

\begin{table*}[t]
\centering
\resizebox{0.9\textwidth}{!}{ 
\begin{tabular}{|l|c|c|c|c|c|c|c|c|c|} \hline  

  & \multicolumn{2}{c|}{\textbf{Ours}} 
  & \textbf{Amortized} 
  & \multicolumn{3}{c|}{\textbf{Model-Agnostic}} 
  & \multicolumn{2}{c|}{\textbf{Model-Specific}} 
  & \textbf{Other} \\ \hline
  & \textbf{FEX-50k} & \textbf{FEX-1.3M} & \textbf{FastSHAP} 
  & \textbf{RISE} & \textbf{IG} & \textbf{GradSHAP} 
  & \textbf{GradCAM} & \textbf{AttLRP} & \textbf{Random} \\ \hline  
 
 \textbf{Positive AUC $\downarrow$} & 0.3573 & \textbf{0.3221} & 0.4591 & 0.5040 & 0.4276 & 0.4599 & 0.5539 & 0.3652 & 0.6350 \\ \hline  
 \textbf{Negative AUC $\uparrow$} & 0.6892 & \textbf{0.7296} & 0.7084 & 0.7229 & 0.7216 & 0.7067 & 0.5546 & 0.7092 & 0.5790 \\ \hline  
 \textbf{Pixel Acc $\uparrow$} & 0.7862 & \textbf{0.8172} & 0.7674 & 0.5022 & 0.5643 & 0.7812 & 0.6786 & 0.8162 & 0.5064 \\ \hline  
 \textbf{mAP $\uparrow$} & 0.6714 & \textbf{0.8939} & 0.6749 & 0.5281 & 0.6135 & 0.6886 & 0.7311 & 0.8590 & 0.5050 \\ \hline  
 \textbf{mIoU $\uparrow$} & 0.4685 & \textbf{0.6587} & 0.4811 & 0.3022 & 0.3714 & 0.4958 & 0.4458 & 0.6517 & 0.3235 \\ \hline 
\end{tabular}
}
\caption{Quantitative evaluation of explanation methods on the image classification task. Positive AUC and Negative AUC are evaluated on ImageNet dataset , while Pixel Accuracy (Pixel Acc), mean Average Precision (mAP), and mean Intersection over Union (mIoU) are reported on the image segamentation dataset.}

\label{tab:result}
\end{table*}

\begin{table*}[t]
\centering
\resizebox{0.7\textwidth}{!}{%
\begin{tabular}{|l|c|l|c|c|c|c|c|}
\hline
 & \textbf{FEX}  &\textbf{FastSHAP} & \textbf{RISE}  & \textbf{IG} & \textbf{GradSHAP} & \textbf{GradCAM} & \textbf{AttLRP} \\ \hline
time (seconds) & \textbf{7.0}&11.6 & 260.2 & 311.9& 313.2& 14.9 & 106.8 \\ \hline
memory (GB) & 2.0  &\textbf{1.2}& 15.9  & 24.5& 7.1& 1.9 & 2.0 \\ \hline
time $\times$ memory & 14.0 &\textbf{13.9}& 4,137.2 & 7641.6& 2,223.7& 28.3 & 213.6 \\ \hline
\end{tabular}
\label{tab:resource}
}
\caption{Experiments on the inference cost for explaining 1000 image predictions of a pretrained ViT model. All experiments are conducted on the same machine with 8 CPU cores and 1 Nvidia A100 GPU.}
\label{tab:infer_speed}
\end{table*}

\begin{table*}[th]
\centering
\resizebox{0.9\textwidth}{!}{%
\begin{tabular}{|c|ccc|cc|ccc|cc|}
\hline
 & \multicolumn{3}{c|}{Trajectory Length $s$} & \multicolumn{2}{c|}{Training Data Size} & \multicolumn{3}{c|}{KL Coefficient $l_{kl}$} & \multicolumn{2}{c|}{Trainable $g(x)$} \\ \hline
 & $s=1$ & $s=5$ & $s=10$ & FEX-50k & FEX-1.3M & $l_{kl}=0$ & $l_{kl}=0.5$ & $l_{kl}=1$ & UNet & ViT \\ \hline
Positive AUC $\downarrow$ & 0.3383 & \textbf{0.3221} & 0.3222 & 0.3573 & \textbf{0.3221} & 0.3897 & 0.3294 & \textbf{0.3221} & 0.3352 & \textbf{0.3221} \\ \hline
Negative AUC $\uparrow$ & 0.6977 & \textbf{0.7296} & 0.7282 & 0.6892 & \textbf{0.7296} & 0.6616 & 0.7185 & \textbf{0.7296} & 0.7130 & \textbf{0.7296} \\ \hline
\end{tabular}
}
\caption{Combined performance comparison across trajectory length $s$, training data size, KL-divergence coefficient $l_{kl}$, and trainable $g(x)$.}
\label{tab:combined}
\end{table*}

\subsection{Baselines} 
For the image classification task, we evaluate our proposed method against six baseline approaches, encompassing model-specific, model-agnostic, and amortized explanation techniques. The model-specific baselines include GradCAM, where we use the last hidden state as the target feature map, and AttLRP , where the default configurations from the original work are utilized. The model-agnostic baselines include GradSHAP, RISE and Integrated Gradients (IG). They require a number of queries ($K$) to the prediction model. In our experiments, $K$s are set to 100 for all model-agnostic baselines.  For the amortized methods, we include FastSHAP, where the explainer is implemented as a U-Net generating a 14×14 heatmap and is trained on 50,000 ImageNet samples (We are not able to train FastSHAP on the full ImageNet dataset because it's extremely slow).

For the text classification task, due to the discrete nature of text tokens, FastSHAP, IG and GradSHAP are not directly applicable. Hence, we only compare our method with RISE, GradCAM and AttLRP, with random attribuiton as a reference.

\subsection{Metrics and Results}
Figure~\ref{fig:cvexamples} illustrates qualitative comparisons of various explanation methods for the image classification task. Our approach achieves comparable visual quality to model-specific methods such as AttLRP and GradCAM, while significantly surpassing model-agnostic baselines like IG and GradSHAP.

For quantitative evaluation, we follow the strategies outlined in \cite{chefer2021transformer}. For image classification, attribution performance is assessed using the area under the curve (AUC) of prediction accuracies, computed by progressively masking features \textit{\textbf{(from 0\% masked to 100\% masked)}} based on their attributed importance. Positive AUC is calculated by masking the most important features first, whereas Negative AUC begins with the least important features. These evaluations are conducted on a randomly selected subset of 5,000 images from the ImageNet validation set. 

To further assess the quality of explanations, we use an annotated image segmentation dataset \cite{guillaumin2014imagenet} comprising 4,276 images across 445 categories. Segmentation labels serve as ground truth for attribution scores in the classification task. Performance is evaluated using proxy metrics: pixel accuracy, which measures the proportion of the most important patches falling within segmentation boxes; mean intersection over union (mIoU), quantifying overlap between segmentation boxes and features with above-average scores; and mean average precision (mAP), representing the area under the precision-recall curve with attribution scores as predictions. 

Results for FEX and FastSHAP are averaged over three trained explainers, RISE, IG, GradSHAP, and Random are averaged over three runs, while only one run for GradCAM and AttLRP as they are deterministic. Table~\ref{tab:result} demonstrate the superior performance of our proposed FEX method compared to other baselines.

For text classification, we adopt the ERASER benchmark \cite{deyoung2019eraser} and evaluate sentiment predictions on the Movie Reviews dataset \cite{zaidan2008modeling}. Explanations are evaluated by plotting F1 score curves as text tokens are progressively inserted based on their attributed importance. As shown in Figure~\ref{fig:text_results}, our method also achieves better performance on the text classification task.

In terms of efficiency, fine-tuning ViT on 1.3M ImageNet samples for a single epoch takes approximately 5 hours on a single A100 GPU, and one epoch is generally sufficient to achieve high-quality attribution scores. Notably, inference with FEX is lightweight, requiring only a single forward pass of $g(\boldsymbol{x})$. According to Table~\ref{tab:infer_speed}, our framework achieves the same level of inference efficiency as FastSHAP. It reduces inference time by over 97\% and memory usage by 70\% compared to traditional model-agnostic approaches (RISE, IG and GradSHAP).

\section{Ablation Study}

\paragraph{Effect of Trajectory Size}
The trajectory size sampled from the policy can impact its optimization. While longer trajectories can provide richer information for policy learning, computational constraints limit their feasibility during training. Striking a balance between trajectory size and computational efficiency is thus critical. As presented in Table~\ref{tab:combined}, performance improves when the trajectory length increases from $s = 1$ to $s = 5$; however, it saturates when extending the trajectory further to $s = 10$. These results indicate that sampling excessively long trajectories is unnecessary.

\paragraph{Effect of Training Data Size}
Training data size is another crucial factor in achieving robust performance. The results in Table~\ref{tab:combined} highlight that the explainer trained on 1.3 million ImageNet samples (FEX-1.3M) significantly outperforms the one trained on only 50,000 samples (FEX-50k). This underscores the importance of using a sufficiently large dataset to enhance the explainer's generalization and reliability.

\begin{figure}
    \centering
    \includegraphics[width=0.9\linewidth]{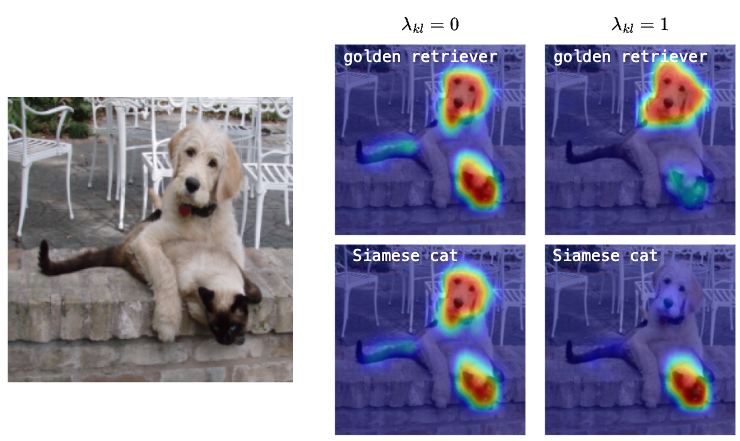}
    \caption{The top two predictions for this image are ``golden retriever" and ``Siamese cat". When $\lambda_{kl}=0$, the explainer cannot differentiate these two classes. While when the KL regularization is introduced, it gains the ability to generalize over different classes.  }
    \label{fig:kl_compare}
\end{figure}

\paragraph{Effect of KL-Divergence Regularization}
The inclusion of KL-divergence regularization enhances the generalizability of the explainer across different classes. As shown in Figure~\ref{fig:kl_compare}, the absence of KL regularization results in a trained explainer that cannot effectively distinguish between classes. Additionally, the results in Table~\ref{tab:combined} indicate that introducing the KL-divergence regularization leads to improved performance.

\paragraph{Impact of $g(\boldsymbol{x})$ Selection}
$g(\boldsymbol{x})$ can be implemented as any neural network that takes $\boldsymbol{x}$ as input and outputs $\boldsymbol{\lambda} \in [0,1]^N$, making it particularly suitable for scenarios where the prediction model $f$ is treated as a black box. To evaluate the effect of different $g(\boldsymbol{x})$ choices, we compared UNet \cite{ronneberger2015u} and Vision Transformer (ViT) architectures. The results in Table~\ref{tab:combined} indicate no significant differences in performance, suggesting that the specific model structure is less critical as long as its capacity (e.g., parameter size) is adequate.

\section{Conclusion}

To address the challenge of balancing general applicability with inference speed in explainable AI (XAI), we proposed FEX framework that bridges the gap between the slow inference speed of model-agnostic methods and the limited applicability of model-specific methods. And unlike amortized approaches, which require existing model-agnostic methods as proxy explainers, our framework has no reliance on any proxy explainers. Experiments demonstrate that our method outperforms other baselines in both explanation quality and inference speed across various metrics.

\clearpage
\section*{Limitations}
Similar to amortized methods, our framework requires training on a large and diverse dataset to achieve better quality, which may pose challenges when data privacy or data acquisition is a concern.  A potential mitigation strategy is to train the explainer jointly with the predictor. This approach not only facilitates explainability for any prediction model but also ensures alignment between the explainer's domain and the predictor's application domain.

\section*{Broader Impact} \label{sec:impact}
Our framework enhances transparency and trust in AI, crucial for applications in sectors like healthcare. It aids debugging and bias identification, supporting ethical AI use and regulatory compliance. However, risks include potential oversimplification of explanations and exposure of proprietary model details. Addressing these challenges is key to maximizing positive impact.

\bibliographystyle{named}
\bibliography{ref}

\end{document}